# Prediction of Bottleneck Points for Manipulation Planning in Cluttered Environment using a 3D Convolutional Neural Network

Indraneel Patil
*EEE Department*
*BITS Pilani, Dubai Campus, UAE*
*e: f20150141@dubai.bits-pilani.ac.in*

B.K. Rout
*Mechanical Department*
*BITS Pilani, Pilani Campus, RJ, India*
*e: rout@pilani.bits-pilani.ac.in*

V.Kalaichelvi
*EEE Department*
*BITS Pilani, Dubai Campus, UAE*
*e: kalaichelvi@dubai.bits-pilani.ac.in*

***Abstract*— Latest research in industrial robotics is aimed at making human robot collaboration possible seamlessly. For this purpose, industrial robots are expected to work on the fly in unstructured and cluttered environments and hence the subject of perception driven motion planning plays a vital role. Sampling based motion planners are proven to be the most effective for such high dimensional planning problems with real time constraints. Unluckily random stochastic samplers suffer from the phenomenon of 'narrow passages' or bottleneck regions which need targeted sampling to improve their convergence rate. Also identifying these bottleneck regions in a diverse set of planning problems is a challenge. In this paper an attempt has been made to address these two problems by designing an intelligent 'bottleneck guided' heuristic for a Rapidly Exploring Random Tree Star (RRT*) planner which is based on relevant context extracted from the planning scenario using a 3D Convolutional Neural Network and it is also proven that the proposed technique generalizes to unseen problem instances. This paper benchmarks the technique (bottleneck guided RRT*) against a 10% Goal biased RRT* planner, shows significant improvement in planning time and memory requirement and uses ABB 1410 industrial manipulator as a platform for implantation and validation of the results.**

## I. Introduction

Manipulation planning can be defined as the kinodynamic planning problem of the end effector of a robotic arm while interacting with a cluttered environment. Kinodynamic problem is a high dimensional planning problem with simultaneous kinematic and dynamic constraints [1]. In this paper a pseudo real time cartesian planning technique in 3D is presented for robotic manipulators which is a problem with a computational time complexity of NP-hard [2]. Sampling based planners are proven to be highly effective for such a problem as they help in achieving a trade off between optimality and computational efficiency by settling for solutions that are just good enough. Sampling domain of such planners is the reachable workspace of an industrial manipulator. Collaborative industrial robots with autonomous real time manipulation are still infeasible because they lack the 3D spatial understanding of the world around them. Sampling based planners also suffer from the 'narrow passage' problem which are regions where the probability of generating samples is low and act as bottlenecks to the planning problem. These regions usually need targeted sampling and more effort to pass through. Machine learning can be used to address these issues to step closer to truly intelligent planning in it's configuration space. Current work proposes the application of a 3D CNN since Convolutional Neural Networks are known to understand complex spatial features, capable of extracting relevant information from the obstacle environment around the industrial robot to predict points from relevant bottleneck regions and heuristically biasing the search of a sampling based planner towards them. The proposed approach in this paper is presented in Figure 1. This work demonstrates the efficacy on single query motion planners but this approach is motion planner agnostic and can also be used with dynamic planners like Multipartite-RRT and RRTx. Our proposed bottleneck guided RRT* has been proven to have a huge improvement over stochastic samplers especially for manipulation planning where one usually has a good prior idea of the class of environment in the manipulator workspace and the exploration remains probabilistically complete.

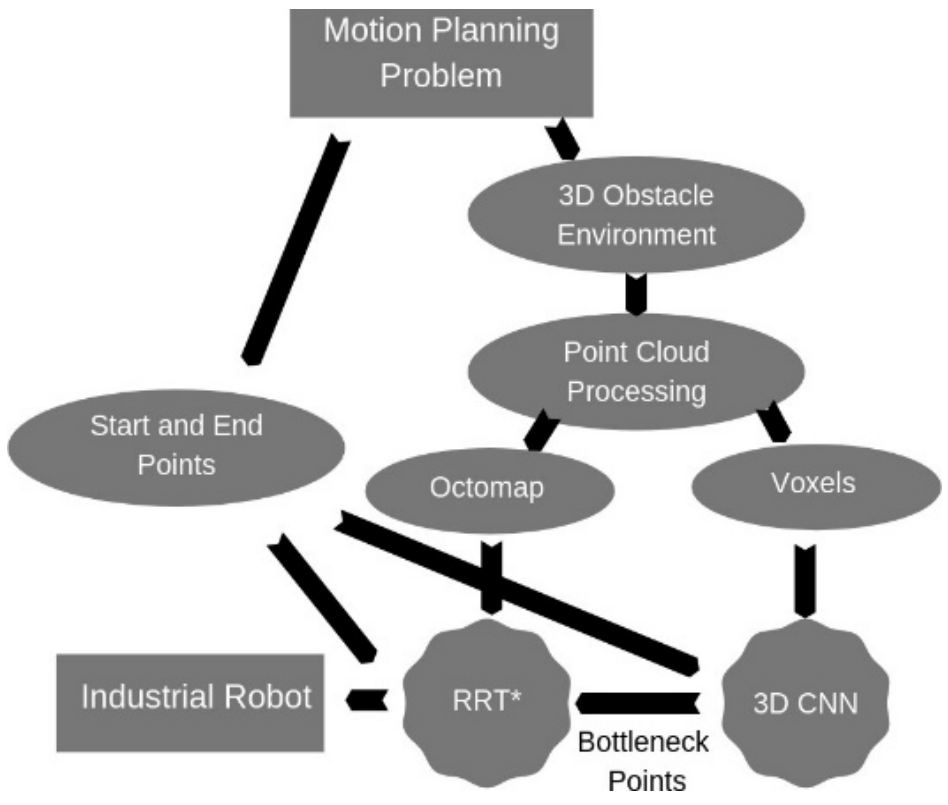

*Figure 1: Flowchart for Bottleneck Guided RRT* Approach.*

In this work, transfer learning technique has been utilized to compensate for the shortage of training data and a pre-trained VoxNet network [3] is reused to predict bottleneck points. A training dataset has been constructed using a sampling based motion planner called as RRT* on challenging hand engineered problems. A Kinect V2 sensor has been used to observe the environment for actual experimentation and results are implemented on ABB 1410 industrial robot.

## II. Background And Motivation

Industrial robotic manipulation involves three complex sub-problems of perception, motion planning and autonomous grasping. This work focusses on the first two sub-problems of enabling perception driven manipulation which is a fundamental pre-requisite for efficient human robot collaborative tasks in cluttered environments. Perception deals with using uncertain sensor measurements to model an accurate estimation of the 3D environment around the manipulator in the form of point clouds, elevation maps, multilevel surface maps, voxels or as octomaps. Out of all the 3D modelling methods voxels allow probabilistic updates of occupied, unknown and free space and allows dimensionality reduction. Octomap also subdivides 3D space into smaller cubic volumes called voxels but it does so based on a tree data structure called as an octree which allows further reduction of the number of nodes needed to be maintained [4].The complex problem of motion planning for industrial robots involves finding a set of configurations which link the start and goal configuration in a cluttered environment without violating a set of constraints like singularity avoidance, physical collisions, self-collision, dynamic constraints and joint limits during the motion. A common approach to tackle this challenge is by splitting it into two stages of global planning and local planning, also called as Hierarchical planning [5]. In the first stage the dynamic constraints have been skipped and a kinematically feasible path has been solved from the start to goal. In the second stage the dynamic constraints are incorporated through path smoothing by various techniques like clothoids [6] , splines [7] , dubin curves [8] and bezier curves [9]. As explained in [10], cubic bezier curves have multiple implementation benefits including continuous curvature along the path, closed form expression for position, computationally efficient to solve for parameters and can easily pass through the knot points as specified by the endpoints of the piece wise linear path specified by global planning.

Sampling based motion planning has been proven to be a more effective option to solve the global planning stage than traditional discrete graph search methods in higher dimensions with more degrees of freedoms in articulated industrial robots [11].They build a dynamic graph online by sampling the configuration space stochastically without the need for explicit modelling of the obstacles offline and converge when a feasible path is found through the graph. The two notable categories in Sampling based planners are of multiple query planners called as Probabilistic Roadmap Technique (PRM) [12] and the single query planners called as the Rapidly Exploring Random Trees[13]are both probabilistically complete hence as the number of samples in the space go to infinity the probability of finding a path (if ones exists) goes to 1. A breakthrough extension of the RRT, called as the Rapidly Exploring Random Trees Star (RRT*) [14] is also asymptotically optimal and always converges to an optimal solution if adequate run time is provided and hence produces much better quality of paths than the RRT algorithm.

In sampling based planners, the order in which samples are chosen to explore the configuration space plays a key role in determining performance of the planner in terms of planning time and memory requirement. This is determined by the expansion method of the planner, which selects new samples in every iteration to grow the graph without violating any internal or external constraints of the industrial robot. There are many drawbacks of uniform sampling which advocate the importance of an intelligent heuristic in this function. Cluttered environments have a lot of narrow passages. A random global sampler has very low probability of generating samples in the narrow region and thus fails to capture the connectivity of the configuration space. As explained in [15], the number of unnecessary nodes added in the tree during sampling grows as the dimensions of the search problem increase and with that the nearest neighbor searches also become more expensive. One of the best ways to tackle the problem is again to develop a computationally intelligent heuristic for your n dimensional problem which will help the tree obtain the goal in fewer samples. Hence there is a lot of research in this area to modify this expansion method to improve performance. A few of the notable heuristics are Goal biasing to greedily connect current configuration to the goal [16], bias sampling towards the stored waypoints from previous solution [17], expansion towards maximum expected utility [18] and steer the tree using shapes of the obstacles in the workspace [19].

Motion planning problems can be quite diverse and there is no 'one size fits all' heuristic solution to all the problems. Also, explicit geometric modelling of the configuration space obstacles becomes infeasible as the dimensions increase. One solution could be to learn different types of motion planning problems and try to replace the motion planner by predicting the entire path from the start to goal. But this approach hurts generalization and essentially overfits to a small group of planning problems [20]. Hence more recently research has turned towards machine learning based approach to extract relevant context from the configuration space obstacles and then design a suitable heuristic for a sampling based motion planner, also coined as adaptive planning. First effort in this direction can be seen in [21] through Randomized Statistical Path Planning (RSPP) which actively adjusts the hyperparameters of the sampling planner according to the motion planning instance but this method still suffers from the 'narrow passage' problem. In [22] , principal component analysis (PCA) is used to detect these narrow passages and then guide the sampler to diffuse through these passages. But the challenging task here is to identify the narrow passage which is hard to generalize and is usually hand engineered. The first work describing different representations of the 3D environment for a learning algorithm including a voxel descriptor was in [23] and they also predicted seeds for trajectory planning. The work discussed in [24] uses a Convolutional Neural Network (CNN) to identify critical regions prior to planning and then leverage this information to help sampling based planners to converge faster. But this work is limited to 2D planar configuration spaces and is not applicable to manipulation planning. The objectives are :

- To demonstrate scene understanding using CNN for manipulation planning.
- First known application of 3D CNN to solve the 'narrow passage' problem and design an adaptive heuristic in motion planning.
- Validation of a multi-input single-output CNN which is easy to train even on a CPU
- Generalizing the proposed network to unseen problem instances using transfer learning.
- The proposed bottleneck guided RRT* achieves 60% improvement in planning time and 80% improvement in memory requirement over traditional 10% Goal Biased RRT*.

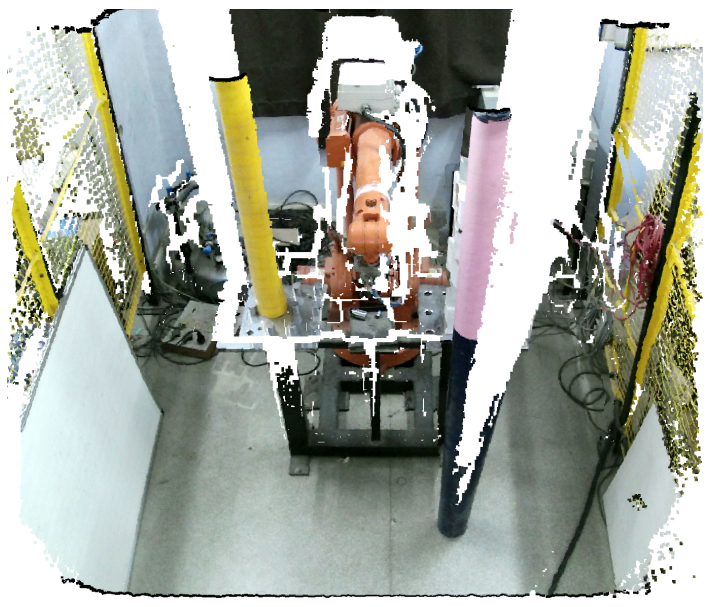

*Figure 2: Original RGB Cloud from the Kinect*

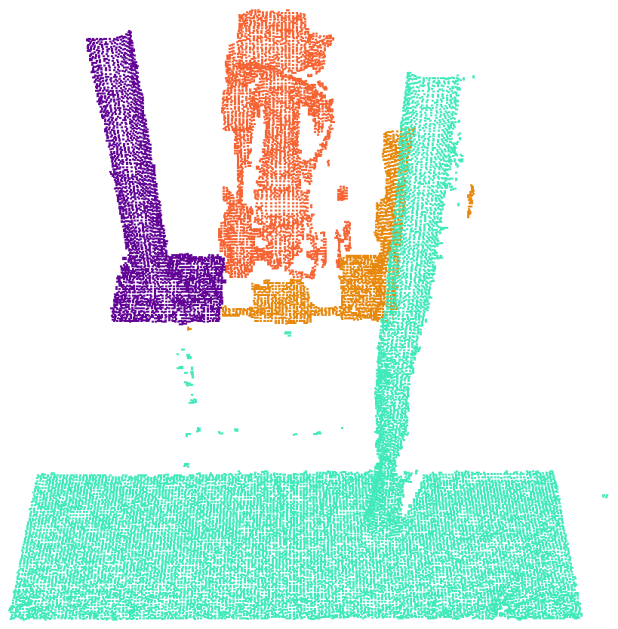

*Figure 3: Colour based Region Growing Segmentation*

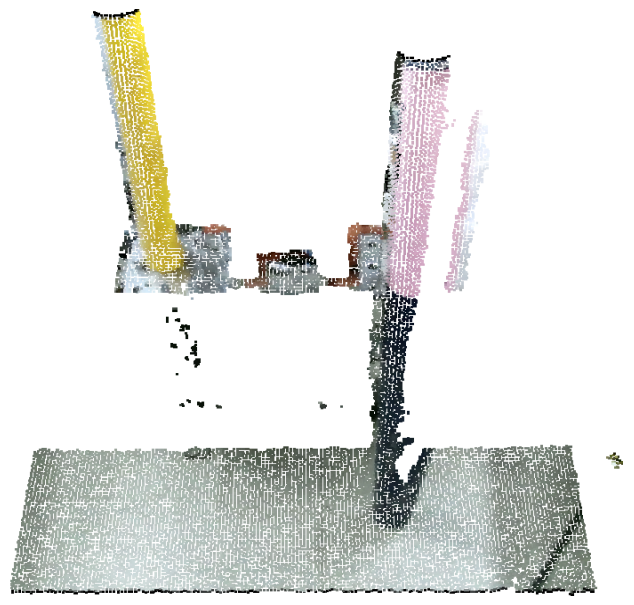

*Figure 4: Output of the Self Identification Process*

## III. 3D Modelling of the Environment

The first task is to obtain a reliable estimate of the environment around the industrial robot using a sensor. Figure 2 gives an RGB point cloud of the 3D scene in the workspace as the starting point of the pipeline. An RGB point cloud consists of algorithmically fusing the RGB information with a point cloud from the depth camera. In this case the 3D modelling problem has been split into two sections of self identification and environment descriptor.

### *A. Self Identification Process*

The Point Cloud Library [25] has been used for all the steps in this section illustrated in Figure 3 and Figure 4. During pre-processing of the cloud a region of interest mask is created using passthrough filters to remove obstacles which are beyond the reachable workspace of the manipulator. Voxel down-sampling is used to reduce the computational burden of processing a full resolution point cloud. A suitably sized Voxel Grid has been used such that the down-sampled cloud still captures the important information of the obstacle environment. In perception based manipulation, there are two possible configurations for the sensor [26]. Firstly in the 'eye in hand' arrangement the sensor is attached on the end affecter and secondly in the 'eye to hand' arrangement the sensor is placed facing the workspace. In this paper the latter approach is chosen so that the viewing angle is large at all times. The first challenge is to compensate the inclined readings from the sensor so that further operations like ray casting are not compromised. Hence a rotational transform of 45 degrees is used followed by a translation to transform the cloud from the kinect frame to the manipulator world frame.

Another challenge that naturally arises in the eye to hand arrangement is that the manipulator needs to be identified and segmented out from the rest of the obstacles. This is the process of self identification. In this work the 'colour based region growing segmentation' method is chosen as a tool for self identification. The industrial robot ABB 1410 is bright orange in color which is unique in the obstacle environment. This virtue of the robot can be used to identify and segment the robot from the RGB cloud. Region growing segmentation algorithm tries to merge neighboring clusters with each other based on particular constraint like smoothness, color and so on. Let $A_i = \{p_i \in P\}$ be a set of all available points at the start of segmentation and $R$ be a set of all regions in point cloud after segmentation.

while $\{A\ \}$ is not empty do:

- Label a seed point $\{S_i\}$ from $\{A\ \}$ which is a point with minimum curvature to represent a region $\{R_i\}$ in the point cloud.
- Find nearest neighbors of the current seed point $\{B\ \} \in \Omega(S_i)$ and while $\{P\ \} \in (B_i)$ is not empty do:
    - For current neighbor point $P_j$ ,if $P_j \in A$ and $\left\|colour(S_i), colour(P_j)\right\| \leq threshold_1$ do :
        - Add $P_j to \{R_i\}$
        - Remove $P_j$ from $\{A\ \}$
- For neighbouring regions of current region $\{R_i\}$, if $\left\|colour(R_i), colour(R_{neighbour})\right\| \leq threshold_2$ then merge to form a bigger region.
- If $colour(R_i)$ is close to orange then delete the region from $R$.

### *B. Environment Descriptor*

The segmented point cloud is used for two use cases. First use case is for collision checking with our motion planner and the second one for scene understanding followed by feature extraction using a 3D CNN. Due to all the competitive advantages along with the computational efficiency, Octomap is used for probabilistically converting the point cloud to each node in an Octree (in the first use case) for collision checking. Octree is defined as a hierarchical data structure for spatial subdivision in 3D where each node in the octree represents the space contained in a cubic volume called a voxel. In this application, root of the octree is the entire reachable workspace of the manipulator. The octree recursively branches into child nodes in areas where obstacles or other objects of interest are present until it reaches leaf nodes which contain the spatial information on obstacles describing the environment around the manipulator.

In the second use case a suitable environment descriptor is needed which can be used as an input to the 3D CNN. In this work a voxel descriptor is used to model the environment directly from sensor data in the form of point cloud. Given a point cloud a 3D grid system of fixed dimensions can be constructed called voxels where value of each voxel is the occupancy probability in that space. Value of each voxel cell is scaled between -1 to 1. 3D ray tracing is applied to find number of hits and pass throughs for each voxel. Each point (x,y,z) in the point cloud is mapped to discrete voxel coordinates (i,j,k). Occupancy grids allows us to efficiently estimate free, occupied and unknown space in the environment of the robot. The main parameters to choose are the origin, orientation and the resolution of the voxel grid in space. A voxel grid is defined with 32 voxels across each dimension where each voxel is a cube with side 0.1m. The idea is to preserve important features in the point cloud and still maintain a computationally efficient size of the voxel grid. So these are hyperparameters which can be tuned for different applications.

## IV. PREDICTING BOTTLENECK POINTS

As proven in [27], the narrow passages in cluttered environments result in bottleneck regions in planning problems as number of samples required here are very high and hence uniform sampling based planners cannot find a path through such passages. Hence in this section such bottleneck points are predicted from narrow passages where the probability of generating samples is low but still need focused sampling in order to find a feasible motion plan. For any motion planning environment there can be many such bottleneck points but only the ones relevant to a planning query are predicted by also spatially considering the start and goal points in the learning model. Thus the learning model has two inputs, the endpoints and the environment descriptor and the outputs are relevant bottleneck points. This section consists of two parts, data generation and network architecture.

### *A. Data Generation and Transfer Learning*

To learn bottleneck points in the supervised learning approach, ideally a huge training and testing data is needed. Due to lack of availability of an open source dataset for the learning task with two inputs and one output, transfer learning from a similar task is chosen and the weights are partially reused in this application. Transfer Learning is defined as improvement of learning in a new task through the transfer of knowledge from a related task that has already been learned [28]. Usually it is the job of the human to provide a mapping from characteristics of one task to those of the other based on correspondences. In this application it is important for the convolutional layers to learn to extract 3D complex features from Voxelized environment descriptor. Hence this knowledge is leveraged from some other task. The VoxNet is used as the original network pre-trained on the Sydney Urban Objects Dataset with 631 urban objects, dataset size of 5261 and 14 class labels [3] . Hence we are confident that this network has the required knowledge to extract the 3D features required for our application.

To create our own manually labelled dataset, a standard 10% goal biased RRT* sampling based planner is used on 200 complex hand engineered planning problems to find first feasible path. It is observed that longest piecewise linear trajectory generated by RRT* had three turns and hence it is assumed that three bottleneck points are enough to capture complexity of any motion planning query. Inspired by the family of planners that use workspace hints to generate samples close to obstacles in order to identify bottleneck points in narrow passages, [19] three points along the first feasible trajectory generated by RRT* which have the lowest clearance from obstacles are declared as training labels for each of the planning problems. Then the knowledge from the original network is transferred to the present task by freezing the weights on the convolutional layers. Thereafter the fully connected layers are replaced with two new dense layers with random initialization and this network is retrained with the manually labelled dataset for the bottleneck prediction task. For testing and training datasets during transfer learning a 1:9 ratio was used.

### *B. Network Architecture and Training*

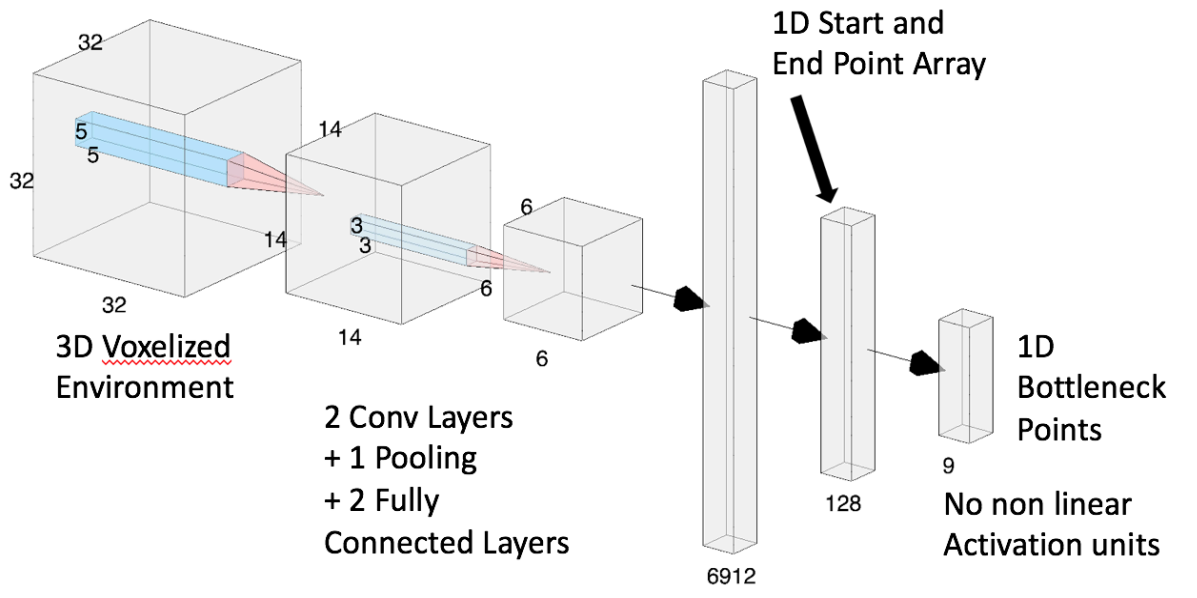

*Figure 5: Multi Input Single Output CNN architecture.*

The 3D CNN architecture is a modified version of the VoxNet network. The network consists of only 2 convolutional layers, one pooling layer and two fully connected layers. This is the simplest CNN for processing 3D data with only 921736 parameters, most of them from the input to the first dense layer and can be trained even on a CPU. The inference is merely a feed forward pass of the network and can be performed efficiently in time intensive motion planning scenarios. The network can extract enough 3D features from the obstacle environment for the motion planning application. There are three main modifications that this paper proposes to the VoxNet for the task. First the network is turned into a Multi Input Single Output CNN where the input to the convolutional layers is the voxelized grid and the start and goal points are an input to the last dense/fully connected layer. Secondly since the output is not a class label but is a finite value, the softmax nonlinearity is removed in the last layer for the prediction task. The last layer has 9 units to predict x,y,z coordinates of three bottleneck points in the environment around the robot as shown in Figure 5.

The input layer accepts a voxel grid of 32×32×32 dimensions. The convolutional layers accept four dimensional input volumes with three spatial and fourth dimension of multiple feature maps. ReLu non-linearity is applied on their outputs. Pooling layer with kernel dimensions of 2×2×2 replace non overlapping blocks of voxel with their maximum. The last fully connected layer combines the extracted 3D features and the endpoints of the plan to predict three relevant bottleneck points. Both the networks are implemented in Tensorflow and training parameters for both are similar except the batch size. Both networks are trained using dropout regularization to reduce overfitting. Adam optimizer with a learning rate of 0.001 is chosen to train the network. The VoxNet was trained for about 20 epochs with softmax cross entropy loss as recommended in [3] whereas the new network was trained for about 170 epochs until the mean squared error loss (MSEL) converges. Lasagne Batch Iterator is used to feed data to Tensorflow feed dictionary in batches which automatically shuffles the dataset. On an average the entire training process took approximately 3 hours on a regular CPU with 2.7GHz Intel Core i5 and 8GB RAM. The manually labelled dataset and the networks are available on Github.

## V. Bottleneck Guided RRT*

This section is dedicated to explaining the new sampling based planner which leverages the predictions and performs a more focused search in the relevant bottleneck regions to converge to a solution faster. The aim is to design a heuristic which uses available predictions from the previous section and improves the visibility of bottleneck or narrow passages.

### *A. Reachability Analysis of Manipulator*

The first requirement for any sampling based planner is a sampling domain or the configuration space $X$. In this case the sampling domain is the same as the reachable workspace of the ABB 1410 industrial manipulator. The OpenRAVE kinematic reachability module [29] is used to find the reachable workspace of the manipulator. After defining the required URDF (Universal Robot Description Format) and collada files of the manipulator which are available on Github the reachability module first analytically finds the inverse kinematics solver for the robot. Then runs an iterative algorithm, which discretizes a cuboidal box around the manipulator into voxels and then tries to find an inverse kinematics solution for each voxel and then adds them to the reachable workspace plot of about 72,46,479 x,y,z points. The rear view and bottom view of the workspace are presented in Figure 6.

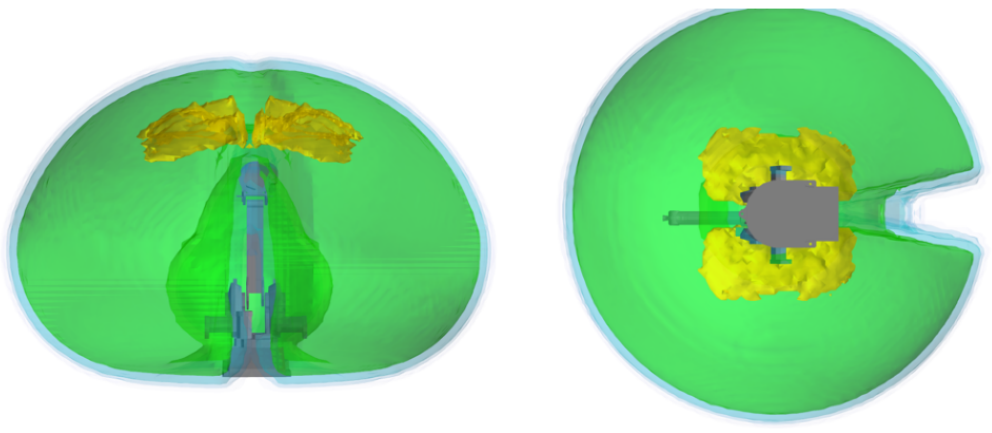

*Figure 6: The Reachable Workspace of the ABB 1410 Industrial Robot.*

### *B. Basic RRT* Algorithm*

RRT* is an incremental sampling based algorithm [14] used in the global planning stage which tries to find a solution by dynamically growing a tree $T$ in $X$ with shortest paths from the start point. Let $X_{free}$ be obstacle free region in configuration space $X$. The tree consists of vertices $V$ which are sampled from the obstacle free space $X_{free}$ and edges $E$ that connect these vertices with each other (ChooseTarget function). The tree keeps growing until both start point $x_{start}$ and goal point $x_{goal}$ are a part of the same tree. In each incremental iteration the tree is grown only by a fixed step size $d$ (Extend function). In RRT* during each incremental step tree is grown towards $x_{new}$ which is a candidate point from $X_{free}$ such that it reduces the overall 'cost to goal' by considering multiple parent points from $V$ for $x_{new}$ (ChooseParent function). Tree is also rewired in each step after this extension in each iteration to achieve the same objective. In our application, cost is defined as Euclidean distance between two vertices in the tree. In the current work the k nearest version of RRT* is implemented which uses the nearest neighbors concept to find multiple parent candidates (Nearest function). We observed that in each step considering three nearest neighbors of the $V_{nearest}$ gives the best results. The expansion method of the planner is responsible for selecting $x_{new}$ in each iteration from $X_{free}$ and can contain a suitable heuristic to select the right points to be added to $V$ to find a plan faster plus rely less on random exploration. Usually this contains a 10% goal heuristic to enable a more directed search of the planner towards the goal. Hence the expansion method contains two probabilities, $p_{goal}$ and $1 - p_{goal}$ .

**Algorithm 1** Goal Biased RRT* Expansion Method

**Require:** tree $T$ and iterations $K$

**for** i = 1......$K$ do
  $x \leftarrow$ function **ChooseTarget**
    p = UniformRandom in [0.0 .. 1.0]
    **if** $p \leq p_{goal}$ then $x = x_{goal}$ **else** $x = x_{rand}$
  $x_{near} \leftarrow$ **Nearest**$(x, V)$
  $x_{min} \leftarrow$ **ChooseParent**$(x, x_{near}, V, E)$
  **if** $\|x_\{min\} - x\|^2 \leq d$ then $x = x_{new}$ **else**
  $x_{new} \leftarrow$ **Extend** $x_{min}$ towards $x$ by a fixed step $d$
  **if** **CollisionFree**$(x_{new})$ then
    $T$.AddVertex$(x_{new})$, $T$.AddEdge$(x_{new}, x_{min})$
    $T$.Rewire$(T, E, x_{new}, x_{min})$
**return** $T$

### *C. New Expansion Method Heuristic*

The RRT* expansion algorithm is redefined to bias sampling towards the bottleneck points as well as preserve the goal bias and the random search of the RRT* algorithm. Now there are three probabilities in the expansion method with $p_1$ for where the goal point is chosen as target, $p_2$ where tree is grown towards bottleneck point and $1 - p_1 - p_2$ where a uniform random point is chosen. This search is a bottleneck guided search which invests more time in exploring the difficult regions than expanding into empty space and helps in converging to the goal much faster than the basic RRT* search. Apart from the ChooseTarget routine in the expansion method, bottleneck guided RRT* works the same way as basic RRT* and preserves the properties of asymptotic optimality and probabilistic completeness but gives much better results in terms of planning time and tree size. We recommend $p_1$ to be 0.2 and $p_2$ to be 0.4 for a cluttered obstacle scene.

**Algorithm 2** Bottleneck guided RRT* Expansion Method

$x \leftarrow$ function **ChooseTarget**
  $p$ = UniformRandom in [0.0 .. 1.0]
  **if** $p \leq p_1$ then $x = x_{goal}$
  **else if** $p > p_1$ and $p \leq p_2$then $x = x_{bottleneck}$
  **else** $x = x_{rand}$

## VI. Implementation and Performance Analysis

Both the 10% goal biased RRT* and bottleneck guided RRT* (b-RRT*) are implemented in C++ as the global planners to solve the kinematic constraints. Mersenne Twister pseudo random number generator was used to sample random points from $X$, the Flexible Collision Library (FCL) for collision checking and proximity queries with the Octomap [30], bezier curves are used as local planners to meet the dynamic constraints and ROS is used for sensor data acquisition and visualization. Planning time till first feasible path and tree size are used as comparative metrics in this analysis. ABB 1410 robot along with IRC5 controller is used as a platform for the implementation of the proposed approach.

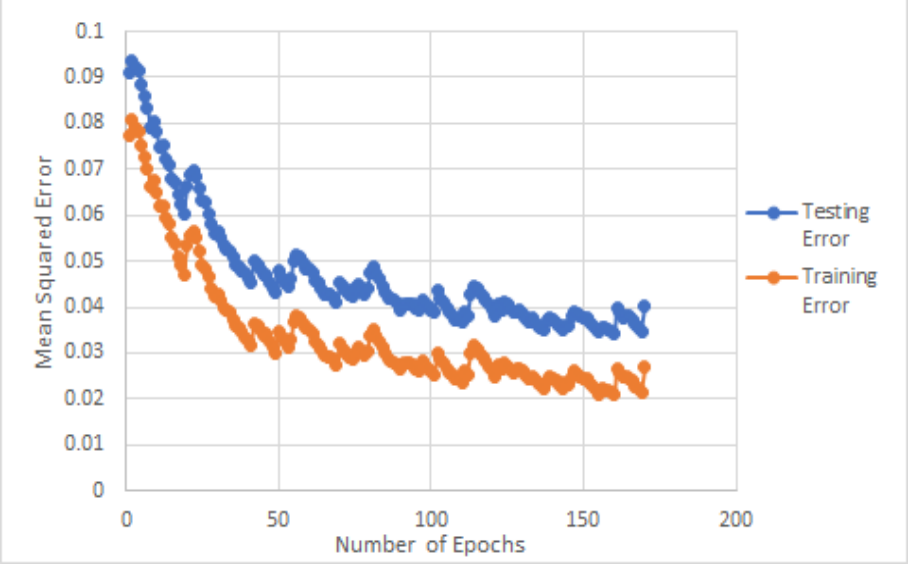

*Figure 7: Learning Curves in Bottleneck Prediction Task*

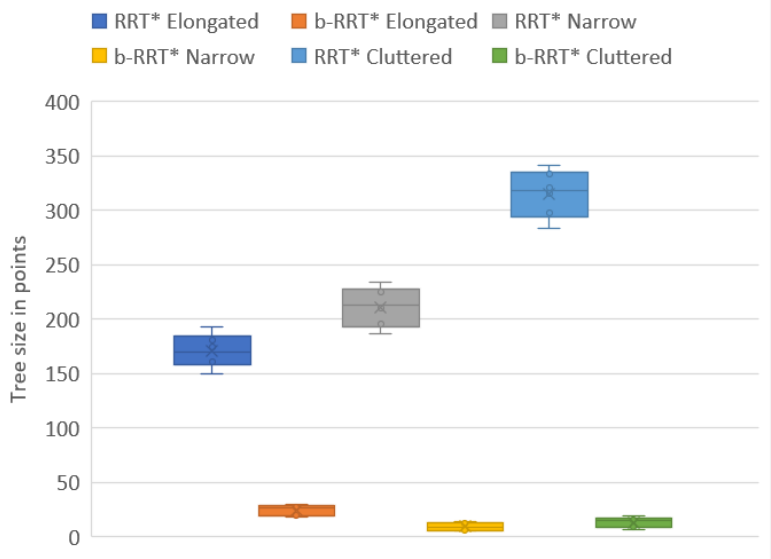

*Figure 8: Tree size comparison of RRT* and b-RRT*.*

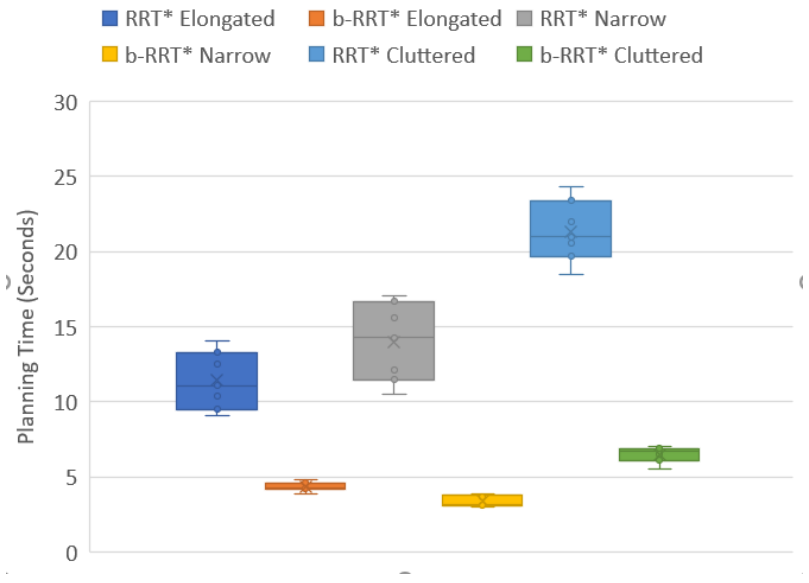

*Figure 9: Planning time comparison of RRT* and b-RRT*.*

The IKFAST tool is used to convert cartesian space trajectories to joint space [31] and the ABB ROS socket connection to download the trajectory to the controller and then play it on the manipulator. During the motion the orientation of end affecter is constant.The planners are compared on three hand engineered realistic environments of elongated human like obstacles, narrow circular spaces and short cluttered scenes as demonstrated in Figure 10, Figure 11 and Figure 12. Our results suggest that bottleneck guided RRT* outperforms a 10% Goal biased RRT* as the complexity of environment increases where the narrow passages dominate the scene. Since a planning time limit was not set the success rate is always 100%. The comparison over 20 cycles by both planners on all three environments is in Figure 8 and Figure 9. On elongated human like environments tree size average showed an improvement of 87% and the planning time improved by 56%. Similar result on narrow circular shaped obstacles of 96% improvement in tree size and 72% improvement in planning time and on cluttered scene with 95% improvement in tree size and 68% improvement in planning time. A simultaneous plot of the training and testing error on the manually labelled dataset presented in Figure 7 suggests that the proposed network is able to learn useful features from the training set and is able to perform well on the unseen problem instances in the test set as well. This proves that the network architecture is well suited for the application in motion planning and can be generalized to real world scenarios.

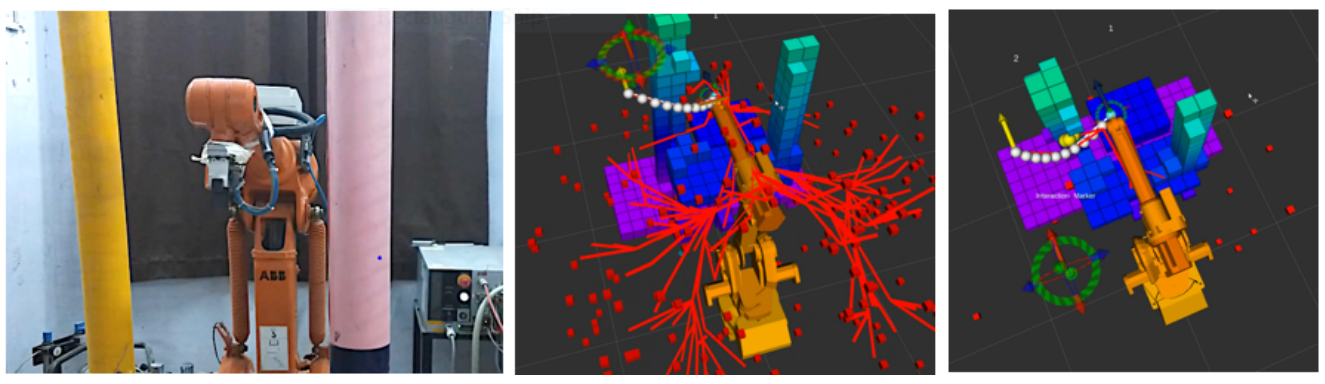

*Figure 10: Elongated human like obstacles: RRT* (12.4s) vs b-RRT* (4.5s)*

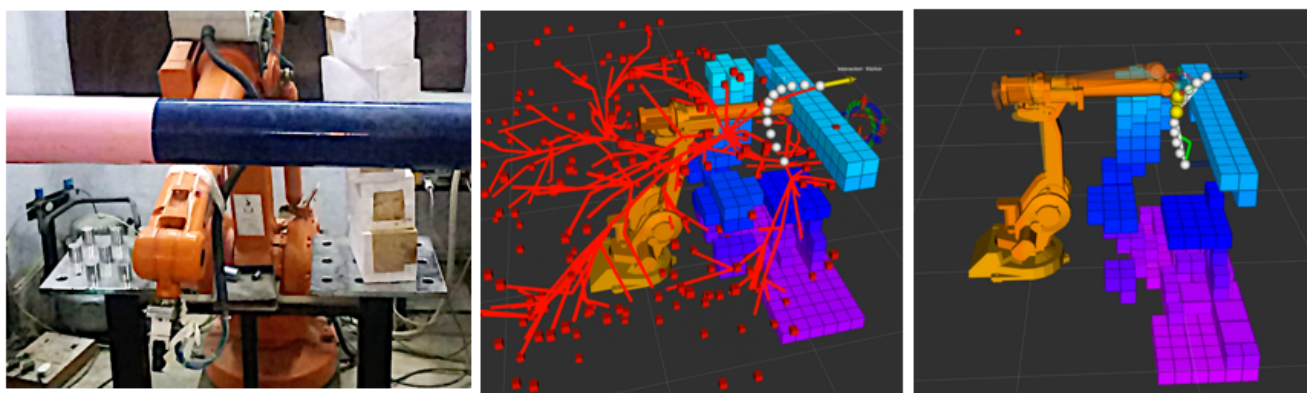

*Figure 11: Narrow circular spaces: RRT*(15.6s) vs b-RRT*(3s)*

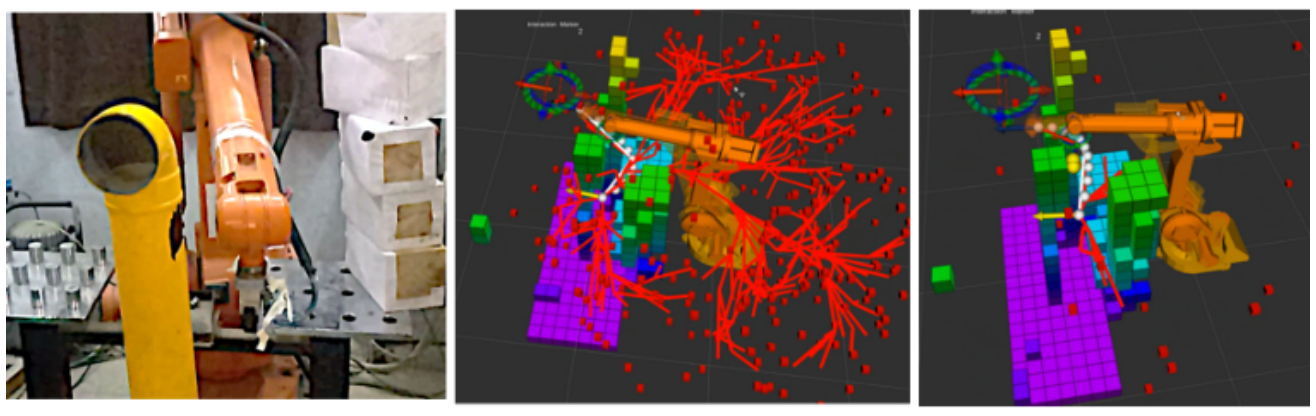

*Figure 12: Cluttered scenes: RRT*(24.7s) vs b-RRT*(5.8s)*

## VII. Conclusion

This paper proposes a new approach in manipulation planning where problems are solved efficiently with an intelligent spatial understanding of the environment and the end points and its performance has been demonstrated with real world experiments. Proposed approach uses a multi input single output 3D CNN capable extracting relevant features from the scene to heuristically guide a sampling based planner even on unseen problem instances using transfer learning. The proposed heuristic helps in achieving a considerable improvement over stochastic search in challenging cluttered environments with experiments on the ABB 1410 industrial robot. Current approach is a step closer to intelligent industrial manipulators capable of working in natural unstructured environments.

Current work has a few limitations that are encountered during implementation. During self identification process the kinematics of the manipulator itself could be used instead of RGB information taken from Kinect sensor. Currently the motion planner is only tested with static obstacles. But this work can also be used with dynamic planners instead like RRTx or Multipartite-RRT with suitable hardware for testing. In the future unsupervised learning could be used to extract context from 3D data without explicit labelling and then this could be used for motion planning

## Acknowledgment

The authors would like to thank the authorities of BITS Pilani, Dubai campus and BITS Pilani, Pilani campus for providing all the encouragement and facilities required during this research work. We are also grateful to Abhishek Chakravarthi and Rohan Saxena for the insightful discussions.